\documentclass[11pt]{article}

\usepackage[T1]{fontenc}
\usepackage[letterpaper,left=0.78in,right=0.78in,top=0.82in,bottom=0.78in]{geometry}
\usepackage{graphicx}
\usepackage{booktabs,array,tabularx}
\usepackage[table]{xcolor}
\usepackage[font=small,labelfont=bf,labelsep=period]{caption}
\usepackage{float}
\usepackage{amsmath,amssymb}
\usepackage{url}
\usepackage[hidelinks]{hyperref}
\usepackage[normalem]{ulem}
\usepackage{textcomp}
\usepackage{fancyhdr}

\definecolor{tableheader}{gray}{0.88}
\definecolor{tablesubheader}{gray}{0.95}

\newcommand{\best}[1]{\textbf{#1}}

\makeatletter
\renewcommand\section{\@startsection{section}{1}{\z@}%
  {1.5ex plus .4ex minus .2ex}{0.8ex plus .2ex}%
  {\normalfont\large\bfseries}}
\renewcommand\subsection{\@startsection{subsection}{2}{\z@}%
  {1.1ex plus .3ex minus .2ex}{-0.45em}%
  {\normalfont\normalsize\bfseries}}
\makeatother

\begin{document}

\begin{center}
{\LARGE\bfseries
Knee3DVLM: Dual-Sequence Full-Volume Vision--Language Modeling for Comprehensive Knee MRI Assessment\par}

\vspace{1.5em}

{\fontsize{11.5}{12.5}\selectfont
Maryam Baizhigitova\textsuperscript{1,2},
Andrew Seohwan Yu\textsuperscript{1,3},
Po-Hao Chen\textsuperscript{1},
Naveen Subhas\textsuperscript{1},\\[0.35em]
Sixu Chen\textsuperscript{1,4},
Xinxin Wang\textsuperscript{1,3},
Kunio Nakamura\textsuperscript{1},
Richard Lartey\textsuperscript{1},\\[0.35em]
Xiaojuan Li\textsuperscript{1,3},
Mingrui Yang\textsuperscript{1}\par}

\vspace{0.9em}

{\fontsize{11.0}{12.5}\selectfont \itshape
\textsuperscript{1}Cleveland Clinic, Cleveland, OH 44106, USA\\
\textsuperscript{2}Cleveland State University, Cleveland, OH 44115, USA\\
\textsuperscript{3}Case Western Reserve University, Cleveland, OH 44106, USA\\
\textsuperscript{4}Kent State University, Kent, OH 44242, USA\\\par}
\end{center}

\vspace{0.2em}
\begin{center}
\begin{minipage}{0.92\textwidth}
Vision-language models (VLMs) are increasingly being applied to three-dimensional medical imaging, with recent systems supporting report generation, visual question answering, localization, and segmentation [1,2]. Their application to knee MRI remains limited, particularly for interpreting the complementary sequences used in clinical practice. The closest existing approach, 3DReasonKnee, processes a single dual-echo steady-state (DESS) series [3], whereas radiologists integrate high-resolution DESS morphology with fluid-sensitive turbo spin-echo (TSE) signal. To address this clinical gap, we introduce Knee3DVLM, a sequence-aware VLM that uses full-volume DESS and TSE inputs to predict 57 anatomically resolved binary diagnostic targets derived from the MRI Osteoarthritis Knee Score (MOAKS) for structured reporting. Three configurations---Knee3DVLM-DESS, Knee3DVLM-TSE, and the fused dual-sequence Knee3DVLM---were evaluated using subject-disjoint Osteoarthritis Initiative partitions [4,5]. In a held-out test cohort of 1,074 examinations, we observed clinically coherent sequence-specific differences: DESS performed better for cartilage and osteophyte abnormalities, whereas fluid-sensitive TSE performed better for bone marrow lesions, meniscal abnormalities, ACL abnormalities, effusion-synovitis, and Hoffa-synovitis. The fused Knee3DVLM achieved 72.98\% average accuracy, 71.17\% balanced accuracy, 78.96\% mean ROC-AUC, and 78.74\% macro ROC-AUC, achieving the highest value among the three configurations on every reported aggregate endpoint. In a secondary multiclass analysis aligned with the released 3DReasonKnee cohort, Knee3DVLM achieved numerically higher accuracy than the strongest reported 3DReasonKnee configuration, Qwen2.5VL-3B, in all five pathology categories. These findings support dual-sequence full-volume modeling that more closely reflects clinical multisequence knee MRI interpretation.

\end{minipage}
\end{center}

\section{Introduction}\label{introduction-1}

Knee osteoarthritis is a highly prevalent joint disease assessed across
the entire joint using complementary imaging planes and pulse sequences
rather than a single view. In routine interpretation, fluid-sensitive
turbo spin-echo (TSE) images commonly serve as the initial survey
because they depict meniscal, cruciate ligament, and marrow
abnormalities, as well as joint effusion, periarticular cysts, and
bursae. Contemporary European Society of Musculoskeletal Radiology
recommendations similarly emphasize multiplanar fluid-sensitive imaging
and systematic evaluation of every joint structure {[}6{]}. This
sequence emphasis is also reflected in the OAI knee protocol: the first
diagnostic acquisition after localization is coronal
intermediate-weighted TSE, while sagittal fat-suppressed TSE provides
the protocol\textquotesingle s distinctive sensitivity to bone marrow
abnormalities and subchondral cysts {[}5{]}. DESS supplies complementary
thin-section, high-resolution depiction of cartilage morphology and
osseous surfaces {[}7,8{]}, but is relatively insensitive to marrow
edema-like signal compared with fluid-sensitive TSE {[}9,5{]}.
Radiologists therefore reconcile TSE and DESS findings across the whole
joint; a DESS-only model cannot fully reproduce this multi-sequence
workflow.

Medical VLMs first demonstrated broad visual question answering and
generative capabilities on predominantly two-dimensional biomedical
figures, radiographs, and other planar images {[}10,11{]}. More recent
systems have extended image-language alignment to volumetric CT and MRI
for report generation, question answering, localization, and
segmentation {[}1,2{]}. Applied to knee MRI specifically, however,
progress remains narrow: existing 3D VLMs generally process a single
image series and are benchmarked on isolated diagnostic questions rather
than comprehensive, multi-sequence whole-joint assessment.

The closest existing system, 3DReasonKnee, narrows this gap only
partially: it introduced regional questions with localization, reasoning
traces, and structured MOAKS severity assessments, but operates on DESS
alone {[}3{]}. More general 3D VLM benchmarks such as M3D-LaMed {[}1{]}
and Med3DVLM {[}2{]} demonstrate efficient full-volume encoding and
language grounding but emphasize CT or single-series inputs and do not
test clinically motivated sequence complementarity.

We close this gap with Knee3DVLM, a sequence-aware framework for
full-volume knee MRI interpretation. Throughout this paper,
\textbf{Knee3DVLM} denotes our paired, dual-sequence model; the
single-sequence ablations are denoted \textbf{Knee3DVLM-\allowbreak{}DESS} and
\textbf{Knee3DVLM-\allowbreak{}TSE}. Knee3DVLM makes three contributions:

\begin{sloppypar}
\begin{enumerate}
\def\labelenumi{\arabic{enumi}.}
\item
  \textbf{Sequence-preserving dual-volume architecture.} Knee3DVLM
  uses independent full-volume DESS and TSE encoders and stream-specific
  tokenizers, retains acquisition identity through learned modality
  embeddings, and performs token-level fusion before projection into the
  language space. This design integrates complementary contrasts without
  collapsing them at the image or encoder-input level.
\item
  \textbf{Sequence-specific and dual-sequence evaluation.} To examine how DESS
  and fluid-sensitive TSE affect VLM performance, we compared DESS-only,
  TSE-only, and fused DESS-TSE configurations on the primary 57-target
  binary task. This analysis tests whether sequence-specific performance
  follows clinically coherent tissue contrasts and whether combining
  both sequences improves overall performance.
\item
  \textbf{Category-wise comparison with 3DReasonKnee.} As a secondary
  benchmark analysis, we aligned the original OAI data with the released
  3DReasonKnee subject, visit, and knee-side identifiers and compared
  category-wise exact-grade average accuracy using the metric reported
  in that study (Table~\ref{tab:3dreasonknee}).
\end{enumerate}
\end{sloppypar}

\section{Related Work}\label{related-work}

\textbf{Semiquantitative Knee MRI Assessment.} The MRI Osteoarthritis
Knee Score (MOAKS) provides a whole-joint framework for semiquantitative
assessment of cartilage damage, bone marrow lesions, osteophytes,
meniscal abnormalities, synovitis, and other structural findings
{[}12{]}. The Osteoarthritis Initiative (OAI) supplies longitudinal MRI
and clinical data at scale {[}4{]}. Sequence selection affects the
conspicuity and measured extent of pathology: fluid-sensitive TSE
imaging offers greater validity and statistical power for bone marrow
lesion assessment, whereas DESS supports high-resolution morphologic
evaluation of cartilage and osseous surfaces {[}9,7{]}.

\textbf{Three-Dimensional Medical Vision-Language Models.} M3D-LaMed
established a general-purpose 3D multimodal model and benchmark spanning
retrieval, captioning, VQA, localization, and segmentation {[}1{]}.
Med3DVLM focused on computationally efficient full-volume encoding
through decomposed convolutions and learned image-language projection
{[}2{]}. These advances motivate full-volume reasoning but do not
directly resolve multi-sequence knee MRI interpretation.

\textbf{Knee MRI Reasoning.} 3DReasonKnee links DESS volumes to
subregion-specific diagnostic questions, bounding boxes,
clinician-authored reasoning, and MOAKS grades {[}3{]}. It is the
closest conceptual predecessor to our study. We retain its emphasis on
anatomically explicit assessment but extend the input space from a
single DESS series to paired DESS and fluid-sensitive TSE, and from
isolated questions to a comprehensive structured whole-joint output.

\section{Methods}\label{methods}

\subsection{Dataset and Cohort.}\label{dataset-and-cohort}

We used the original Osteoarthritis Initiative (OAI) dataset and aligned
it with the cohort manifest released with 3DReasonKnee. Matching was
performed by OAI subject, visit, and knee side. These identifiers were
used to retrieve the corresponding original OAI MOAKS readings and DESS
examination and to pair the fluid-sensitive TSE examination from the
same subject, visit, and knee. Acquisition date provided an additional
consistency check. We defined a study as one labeled subject-visit-knee
examination; consequently, a participant could contribute bilateral
knees and multiple longitudinal visits but could never appear in more
than one data partition. Participants, rather than examinations, were
assigned to fixed training, validation, and test partitions. The
resulting subject-disjoint split contained 2,000 training participants
(75.4\%), 267 validation participants (10.1\%), and 385 test
participants (14.5\%), for a total of 2,652 unique participants. These
participants contributed 5,901 training studies (74.0\%), 837 validation
studies (10.5\%), and 1,232 test studies (15.5\%), or 7,970 labeled
studies overall. A participant-ID audit after cohort assembly and
sequence matching confirmed zero overlap among the partitions; all
knees, visits, and sequences from each participant remained in one
partition. Sequence availability was nearly complete. DESS was available
for 5,899 of 5,901 training studies (99.97\%), 837 of 837 validation
studies (100\%), and 1,231 of 1,232 test studies (99.92\%); TSE was
available for 5,897 of 5,901 (99.93\%), 837 of 837 (100\%), and 1,232 of
1,232 studies (100\%), respectively. Matched DESS-TSE pairs were
available for 5,895 of 5,901 training studies (99.90\%), 837 of 837
validation studies (100\%), and 1,231 of 1,232 test studies (99.92\%).
Each configuration used only studies containing its required sequence or
sequence pair; missing acquisitions were excluded rather than imputed.
For the controlled primary comparison, all three configurations were
evaluated on the same 1,074 fully paired test examinations retained
after preprocessing and prediction matching, preventing differences in
test-set composition from affecting the comparison.

\begin{figure}[H]
\centering
\includegraphics[width=0.86\textwidth]{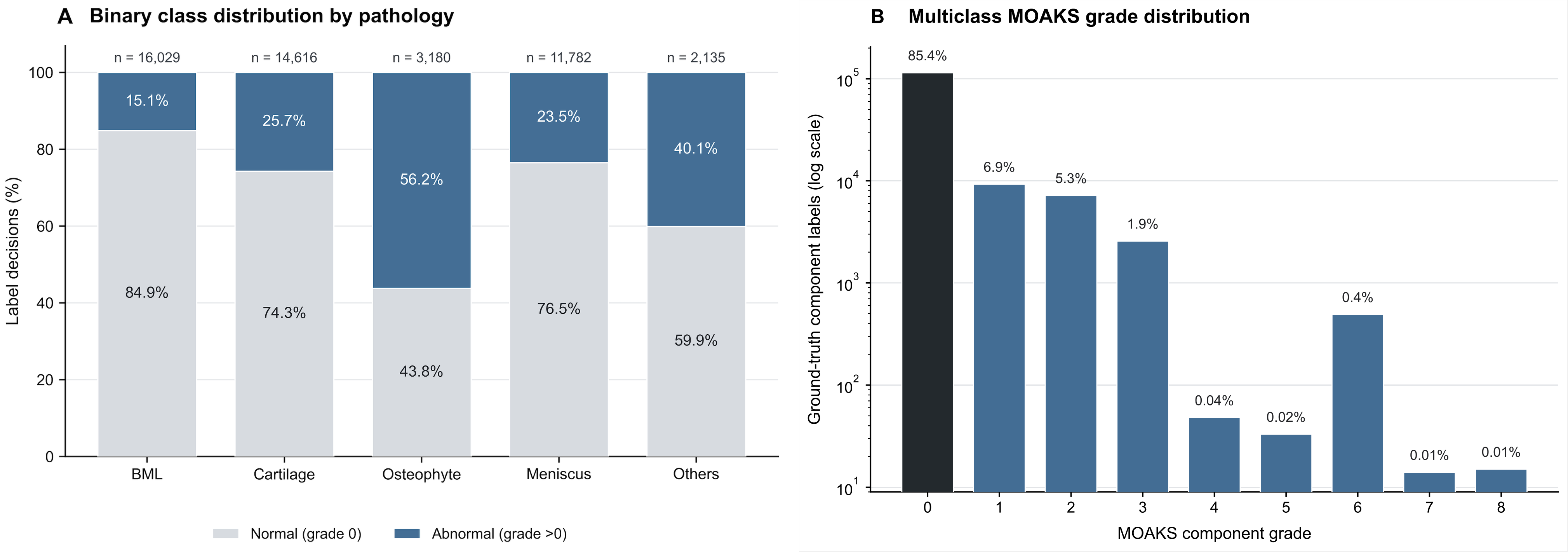}
\caption{Class imbalance in the binary and multiclass evaluation tasks.
Panel A summarizes ground-truth normal-versus-abnormal labels across the
57 targets, grouped by pathology. Panel B shows the pooled ground-truth
grade distribution across all 109 multiclass MOAKS component attributes
on a logarithmic axis, exposing rare grades while retaining the dominant
grade 0 class (85.4\%).}\label{fig:class-imbalance}
\end{figure}

\textbf{Cohort flow and benchmark alignment.} The released 3DReasonKnee
manifest comprised 7,970 labeled subject-visit-knee studies from 2,652
participants. Manifest identifiers were matched to the original OAI
records by subject, visit, and knee side; the corresponding DESS
examination and fluid-sensitive TSE examination from the same knee and
visit were then retrieved. The fixed subject-disjoint test partition
contained 1,232 studies, of which 1,231 had both sequences available.
After preprocessing and requiring matched prediction records from
Knee3DVLM-\allowbreak{}DESS, Knee3DVLM-\allowbreak{}TSE, and Knee3DVLM, 1,074 examinations formed
the common primary test cohort; 157 paired examinations were excluded
during preprocessing or prediction matching. The secondary exact-grade
analysis used the same 3DReasonKnee-aligned identifiers and the original
OAI MOAKS fields, ensuring alignment by subject, visit, and knee side
rather than by generated reports.

\subsection{Target Construction.}\label{target-construction}

As shown in Figure~\ref{fig:class-imbalance}, Panel A, for the primary task, original OAI MOAKS
measurements were collapsed into 57 binary pathology-region targets:
grade 0 was normal, and any valid nonzero grade was abnormal. Each
target therefore represents a specific pathology at a defined anatomical
location or structural feature rather than a generic disease class. Each
$n$ is the total number of evaluable OAI study--target MOAKS annotations
pooled within that group: grade 0 was mapped to normal, a valid nonzero
grade to abnormal, and missing or non-evaluable readings were excluded.
Thus, $n$ counts annotations rather than patients or model predictions.
The
targets cover the medial and lateral compartments, anterior, central,
and posterior subregions where applicable, patellar surfaces, meniscal
segments and extrusion sites, the cruciate ligaments, and whole-joint
inflammatory findings summarized in Table~\ref{tab:binary-schema}.
Missing or non-evaluable MOAKS readings were masked and were never
converted to negative labels.

\begin{table}[H]
\centering
\caption{Binary target schema. Parenthetical counts indicate the number
of binary targets contributed by each anatomical grouping.}\label{tab:binary-schema}
{\fontsize{7.8}{9.2}\selectfont
\renewcommand{\arraystretch}{1.12}
\setlength{\tabcolsep}{4.5pt}
\begin{tabularx}{\textwidth}{@{}
>{\raggedright\arraybackslash}p{1.65cm}
>{\centering\arraybackslash}p{1.15cm}
>{\raggedright\arraybackslash}X@{}}
\toprule
\textbf{Pathology} & \textbf{Targets} & \textbf{Anatomical definition} \\
\midrule
BML & 15 & Femur:\enspace medial/lateral $\times$ anterior/central/posterior (6);\newline
Tibia:\enspace medial/lateral $\times$ anterior/central/posterior (6);\newline
Patella:\enspace medial/lateral (2);\newline
Tibial subspinous region:\enspace one target (1). \\
\addlinespace[3pt]
Cartilage & 14 & Femur:\enspace medial/lateral $\times$ anterior/central/posterior (6);\newline
Tibia:\enspace medial/lateral $\times$ anterior/central/posterior (6);\newline
Patella:\enspace medial/lateral (2). \\
\addlinespace[3pt]
Osteophyte & 12 & Femur:\enspace medial/lateral $\times$ anterior/central/posterior (6);\newline
Tibia:\enspace medial/lateral (2);\newline
Patella:\enspace medial/lateral/superior/inferior (4). \\
\addlinespace[3pt]
Meniscus & 12 & Morphology:\enspace medial/lateral $\times$ anterior horn/body/posterior horn/root (8);\newline
Extrusion:\enspace anterior and body sites on each side (4). \\
\addlinespace[2pt]
ACL & 1 & Whole-ligament anterior cruciate ligament abnormality. \\
PCL & 1 & Whole-ligament posterior cruciate ligament abnormality. \\
Effusion & 1 & Whole-joint effusion-synovitis. \\
Synovitis & 1 & Whole-joint Hoffa-synovitis. \\
\bottomrule
\end{tabularx}}
\end{table}

For the secondary task, we returned to the original OAI MOAKS variables
and retained their exact component grades rather than collapsing them to
normal versus abnormal. This produces 109 grade-bearing attributes
because several anatomical questions contain more than one MOAKS
component: the 15 BML regions each contribute lesion size, lesion count,
and non-cystic lesion percentage (45 attributes); the 14 cartilage
regions each contribute lesion size and depth (28); 12 osteophyte
locations each contribute one grade (12); meniscal morphology and
extrusion contribute 10 attributes; and 11 questions grouped as Others
contribute 14 attributes spanning Hoffa-synovitis, effusion-synovitis,
ACL and PCL features, patellar-tendon signal, bursae, and cysts. Thus,
57 and 109 are not competing class counts: 57 is the prespecified binary
whole-joint output space, whereas 109 is the expanded exact-grade
component space. A schema audit confirmed that all 109 grading
attributes used for the comparison mapped directly to original OAI MOAKS
fields and matched the definitions in the released 3DReasonKnee question
schema (109/109, 100\% coverage). No generated 3DReasonKnee report was
used as ground truth. This exact-grade analysis is secondary because the
component-grade distribution is highly imbalanced.

\subsection{Model Architecture.}\label{model-architecture}

\begin{figure}[H]
\centering
\includegraphics[width=\textwidth]{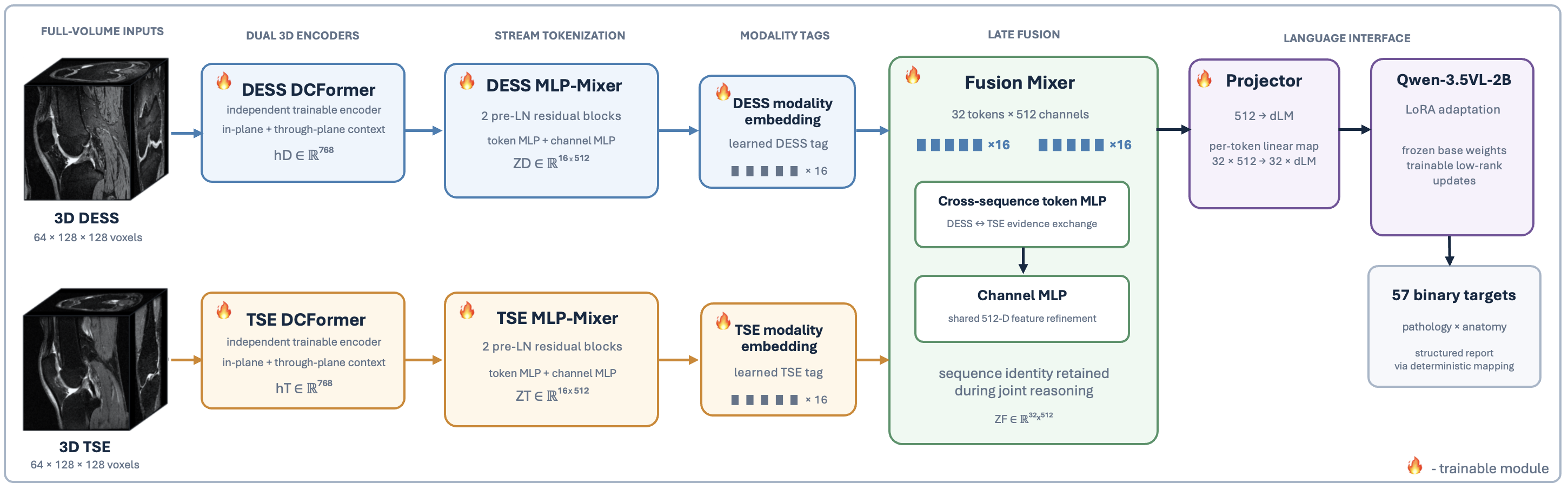}
\caption{Knee3DVLM architecture. Independent trainable DCFormer encoders
map the full DESS and TSE volumes to 768-dimensional representations.
Sequence-specific MLP-Mixers form 16 visual tokens of width 512 per
sequence. Learned modality embeddings preserve sequence identity before
the 32 tokens undergo cross-sequence token mixing and shared channel
refinement. A per-token projector maps the fused representation from
width 512 to the language-model embedding dimension
$d_{\mathrm{LM}}$. The LoRA-adapted language backbone predicts 57
anatomically resolved binary targets, which are converted into a
structured report by deterministic mapping. Flame symbols identify
trainable modules.}\label{fig:architecture}
\end{figure}

Knee3DVLM is organized as a hierarchical late-fusion architecture in
which sequence-specific volumetric representation learning precedes
cross-sequence interaction and language conditioning
(Fig.~\ref{fig:architecture}). Let
$X_D,X_T\in\mathbb{R}^{1\times64\times128\times128}$ denote the paired
DESS and TSE volumes, where the spatial dimensions correspond to depth,
height, and width. For modality $m\in\{D,T\}$, the forward pathway is
\begin{equation}
h_m=f_m(X_m),\qquad Z_m=g_m(h_m),\qquad
\widetilde{Z}_m=Z_m+\mathbf{1}_{16}E_m,
\label{eq:architecture}
\end{equation}
followed by
\begin{equation}
Z_F=q\!\left([\widetilde{Z}_D;\widetilde{Z}_T]\right),\qquad
H_V=p(Z_F).
\label{eq:fusion-projection}
\end{equation}
Here, $f_m$ is the sequence-specific three-dimensional encoder,
$h_m\in\mathbb{R}^{768}$ is its examination-level representation,
$g_m$ is the stream-specific tokenizer, and
$Z_m\in\mathbb{R}^{16\times512}$ contains the resulting visual tokens.
$E_m\in\mathbb{R}^{512}$ is a learned modality embedding that is added
to all 16 tokens through broadcasting by $\mathbf{1}_{16}$, thereby
identifying each token as DESS or TSE. The bracket $[\widetilde{Z}_D;\widetilde{Z}_T]$ denotes
concatenation along the token dimension, $q$ is the cross-sequence
fusion Mixer, and $p$ is the language-space projector. Consequently,
$Z_F\in\mathbb{R}^{32\times512}$ and
$H_V\in\mathbb{R}^{32\times d_{\mathrm{LM}}}$, where
$d_{\mathrm{LM}}$ is the hidden embedding dimension required by the
language model. This decomposition imposes three explicit stages of
representation learning: spatial reasoning within each volume,
contrast-specific compression within each sequence, and joint reasoning
across the paired examination. Cross-sequence interaction therefore
occurs only after each acquisition has formed a complete volumetric
representation.

\textbf{Dual-encoder full-volume representation learning.} Knee3DVLM
receives spatially matched DESS and TSE volumes from the same subject,
visit, and knee. Each sequence is reoriented to a common anatomical
convention, resampled to $64\times128\times128$ voxels (depth $\times$
height $\times$ width), and intensity-standardized independently so that
its native contrast distribution is retained. The volumes are processed
by two independently parameterized DCFormer encoders {[}2{]}, not a
shared visual trunk. Within each encoder, decomposed in-plane and
through-plane operations capture spatial morphology and interslice
context across the complete volume. Global aggregation and a learned
projection produce a 768-dimensional sequence representation,
$h_m\in\mathbb{R}^{768}$ for $m\in\{D,T\}$. The
DESS and TSE encoders are initialized from their corresponding
sequence-specific checkpoints and remain trainable, enabling each branch
to retain its contrast specialization while adapting to the common
57-target whole-joint objective. Because the branches are independently
parameterized, optimization is not forced to represent the markedly
different DESS and TSE intensity distributions with a single set of
visual filters. At the same time, end-to-end training allows gradients
from the shared diagnostic objective to update both encoders, so their
representations become complementary at the point of fusion rather than
remaining fixed feature extractors.

\textbf{Fixed-width stream-specific tokenization.} Each
768-dimensional encoder output is expanded by a separate trainable
MLP-Mixer into 16 tokens of width 512. Each stream contains two
pre-normalized residual Mixer blocks: a token-mixing multilayer
perceptron exchanges information across the 16-token axis, followed by a
channel-mixing multilayer perceptron operating across the 512-dimensional
feature axis. The DESS and TSE tokenizers share no parameters, allowing
the two contrast-specific compression functions to develop independently
before multimodal interaction. This yields
$Z_m\in\mathbb{R}^{16\times512}$ for each sequence. More formally, a
Mixer block applies residual token mixing and residual channel mixing,
\begin{equation}
U=Z+\mathcal{M}_{\mathrm{tok}}(\operatorname{LN}(Z)),\qquad
Z'=U+\mathcal{M}_{\mathrm{ch}}(\operatorname{LN}(U)),
\label{eq:mixer}
\end{equation}
where $\operatorname{LN}$ denotes Layer Normalization,
$\mathcal{M}_{\mathrm{tok}}$ exchanges evidence among learned visual
tokens, and $\mathcal{M}_{\mathrm{ch}}$ transforms their feature
channels. Separate parameters are used at every DESS and TSE
tokenization stage. The resulting learned bottleneck converts the two
full volumes---more than two million input voxels in total---into a
fixed visual budget of 32 tokens while preserving distinct token sets
for the two acquisitions.

\textbf{Modality-aware hierarchical late fusion.} Cross-sequence
interaction is deliberately delayed until both volumes have undergone
independent 3D encoding and within-stream token refinement. A learned
512-dimensional modality embedding is added to every token, making its
DESS or TSE origin explicit. The two streams are concatenated into
$Z\in\mathbb{R}^{32\times512}$ and processed by a fusion Mixer block that
performs token and channel mixing across the complete paired examination.
Thus, token mixing can exchange evidence across sequences, whereas the
retained modality tags allow the fusion module to distinguish the source
contrast of that evidence. The residual fusion pathway also provides a
direct information route from each sequence-specific token set to the
projected representation.
Only after fusion is each token mapped through a single linear projector
$W_{\mathrm{out}}\in\mathbb{R}^{512\times d_{\mathrm{LM}}}$ into the
language-model hidden embedding dimension $d_{\mathrm{LM}}$:
\[
\mathbb{R}^{768}\!\times2\ \longrightarrow\
\mathbb{R}^{16\times512}\!\times2\ \longrightarrow\
\mathbb{R}^{32\times512}\ \longrightarrow\
\mathbb{R}^{32\times d_{\mathrm{LM}}}.
\]
This fixed-width bridge is computationally important: all token and
channel mixing occurs at width 512, independent of the language backbone,
and only the terminal per-token projection scales with
$d_{\mathrm{LM}}$. The visual
interface therefore grows linearly, rather than quadratically, with the
language-model hidden size. In addition, the language model always
receives 32 visual tokens, so the visual contribution to its attention
sequence length is fixed and does not grow with the original volume
dimensions. This separation makes the fusion interface portable across
language backbones with different hidden sizes: only the terminal
projector must conform to the selected language embedding dimension.

\textbf{Architectural contribution.} The resulting hierarchy combines
two independently trained volumetric encoders with sequence-preserving
compression and late representation-level fusion. It avoids premature
contrast entanglement, retains acquisition identity during joint
reasoning, and supplies the language model with a compact paired-volume
representation. Importantly, fusion is performed over learned
three-dimensional examination-level features rather than over isolated
slices or independently generated textual descriptions. The language
backbone consequently receives one jointly optimized representation of
the paired examination, while the visual pathway remains explicitly
aware of which evidence originated from DESS and which originated from
TSE. The same modular interface yields controlled
Knee3DVLM-\allowbreak{}DESS and Knee3DVLM-\allowbreak{}TSE paths by retaining one
encoder-tokenizer stream without changing the language interface,
prediction space, or reporting procedure.

\textbf{Language-conditioned multi-target prediction.} The 32 projected
visual tokens condition a Qwen-3.5VL-2B backbone adapted with trainable
low-rank adapters (LoRA). The optimized pathway comprises both DCFormer
encoders, both stream Mixers, the fusion Mixer, the language projector,
and the LoRA parameters. The pretrained language weights supply the
language prior, whereas LoRA provides a parameter-efficient route for
adapting multimodal reasoning to the structured knee MRI task. Because
the visual encoders, tokenizers, fusion module, projector, and adapters
are optimized jointly, supervision from the complete target set is
propagated through the entire multimodal pathway. A single
examination-level forward pass produces
probabilities for 57 anatomically resolved binary targets derived from
MRI Osteoarthritis Knee Score (MOAKS) readings. Each output is tied to a
defined pathology-location pair rather than a generic whole-knee label.
A deterministic reporting layer then organizes target-level predictions
by pathology and anatomical subregion and constructs the findings,
clinical impression, and overall conclusion. Fixing this final mapping
ensures that comparisons reflect diagnostic predictions rather than
variation in generated phrasing.

\subsection{Training and Inference.}\label{training-and-inference}

All models were trained using only the training partition; validation
data were used for checkpoint selection, and the test set remained
untouched until final evaluation. Volumes were intensity-clipped,
standardized, and resampled to $64 \times 128 \times 128$ voxels before being passed
through the appropriate sequence-specific encoder. The binary task used
class-aware training and target-specific validation thresholds. At
inference, each model generated a continuous probability of abnormality
for every target. A separate decision threshold was selected for each
target using validation predictions only. The threshold maximized
target-level balanced accuracy, with ties resolved in favor of the value
closest to 0.50, and was locked before test evaluation. A test
probability equal to or greater than its validation-derived threshold
was classified as positive.

Continuous probabilities were retained for threshold-independent
evaluation with ROC-AUC. Thresholded predictions were used to
calculate accuracy, sensitivity, specificity, and balanced accuracy.
Missing test labels were excluded separately for each target so
that every metric used only evaluable observations.

Finally, the deterministic reporting layer organized the binary
predictions into pathology-specific findings and anatomical subregions.
Positive predictions were reported as detected abnormalities, while
evaluable negative predictions were represented as normal findings. The
same reporting procedure was applied to all three model configurations,
ensuring that comparisons reflected differences in model predictions
rather than differences in report generation.

\subsection{Evaluation.}\label{evaluation}
For binary target $i$ and examination $j$, $y_{ij}\in\{0,1\}$ denotes
the observed label, $p_{ij}$ the predicted probability of abnormality,
and $\tau_i$ the target-specific decision threshold selected exclusively
on the validation set by maximizing balanced accuracy; ties were
resolved in favor of the threshold closest to 0.50. The locked test
prediction was $\hat y_{ij}=\mathbb{1}[p_{ij}\geq\tau_i]$. Missing and
non-evaluable labels were excluded target by target.

\textbf{Threshold-dependent classification metrics.} For every target,
we calculated ordinary accuracy, sensitivity (recall), specificity,
and balanced accuracy from the test-set confusion matrix:

\[
\mathrm{Acc}_i=\frac{TP_i+TN_i}{N_i},\quad
\mathrm{Sens}_i=\frac{TP_i}{TP_i+FN_i},\quad
\mathrm{Spec}_i=\frac{TN_i}{TN_i+FP_i}.
\]

\[
\mathrm{BA}_i=\tfrac12(\mathrm{Sens}_i+\mathrm{Spec}_i).
\]

Balanced accuracy was the primary endpoint because it assigns equal
weight to abnormal-case sensitivity and normal-case specificity despite
large differences in target prevalence. Average accuracy and every
macro metric were unweighted means across the $K=57$ targets, so targets
with more evaluable observations could not dominate the summary:

\[
\mathrm{Macro}(M)=\frac{1}{K}\sum_{i=1}^{K}M_i,\qquad
\text{Average accuracy}=\frac{1}{K}\sum_{i=1}^{K}\mathrm{Acc}_i.
\]

For target $i$, $\mathrm{ROC\mbox{-}AUC}_i=\Pr(s_i^+>s_i^-)+
\tfrac12\Pr(s_i^+=s_i^-)$, where $s_i^+$ and $s_i^-$ are predicted
scores for randomly selected positive and negative examinations. The
mean ROC-AUC is the evaluable-examination-weighted mean of the
target-level ROC-AUC values,
\[
\mathrm{Mean\ ROC\mbox{-}AUC}
=\frac{\sum_{i=1}^{K}N_i\,\mathrm{ROC\mbox{-}AUC}_i}
{\sum_{i=1}^{K}N_i},
\]
where $N_i$ is the number of evaluable examinations for target $i$.
To ensure that targets with fewer evaluable labels retained equal
influence, we also calculated macro ROC-AUC,
\[
\mathrm{Macro\ ROC\mbox{-}AUC}
=\frac{1}{K}\sum_{i=1}^{K}\mathrm{ROC\mbox{-}AUC}_i,
\]
Mean ROC-AUC gives greater influence to targets with more evaluable
examinations, whereas macro ROC-AUC gives every target equal influence.
Both are reported to distinguish observation-weighted discrimination
from equal-target discrimination under the observed missingness and
imbalance. Targets lacking either class were excluded.

\textbf{Secondary exact-grade evaluation.} To match 3DReasonKnee,
category-wise average accuracy was calculated as the unweighted mean of
attribute-level exact-grade accuracies within each pathology category.
Participant-level confidence intervals and paired significance tests
were not calculated and are acknowledged as limitations.

\section{Results}\label{results}

\subsection{Primary 57-Target Binary
Results.}\label{primary-57-target-binary-results}

Table~\ref{tab:primary-binary} compares the three Knee3DVLM input configurations on the primary
task. Knee3DVLM achieved the highest value among the three
configurations for every reported aggregate endpoint: average
accuracy (72.98\%), balanced accuracy (71.17\%), mean ROC-AUC
(78.96\%), and macro ROC-AUC (78.74\%).
Knee3DVLM-\allowbreak{}DESS ranked second on each aggregate endpoint, while
Knee3DVLM-\allowbreak{}TSE contributed complementary information in several pathology
groups (Table~\ref{tab:pathology-ba}).

\begin{table}[H]
\captionsetup{skip=3pt}
\centering
\caption{Primary binary 57-target performance on the matched
held-out test cohort. Values are presented in percentages. }\label{tab:primary-binary}
{\fontsize{8}{9.4}\selectfont\setlength{\tabcolsep}{3.2pt}
\renewcommand{\arraystretch}{1.12}
\resizebox{\textwidth}{!}{%
\begin{tabular}{lrrrrrrr}
\toprule
& \multicolumn{1}{c}{\textbf{Accuracy}} &
\multicolumn{3}{c}{\textbf{Threshold-dependent classification}} &
\multicolumn{2}{c}{\textbf{Discrimination}} &
\multicolumn{1}{c}{\textbf{Cohort}} \\
\cmidrule(lr){2-2}\cmidrule(lr){3-5}\cmidrule(lr){6-7}\cmidrule(l){8-8}
\textbf{Model} &
\shortstack{\textbf{Average}\\\textbf{accuracy}} &
\shortstack{\textbf{Balanced}\\\textbf{accuracy}} &
\textbf{Sensitivity} &
\textbf{Specificity} &
\shortstack{\textbf{Mean}\\\textbf{ROC-AUC}} &
\shortstack{\textbf{Macro}\\\textbf{ROC-AUC}} &
\shortstack{\textbf{Examinations}\\\textbf{($n$)}} \\
\midrule
Knee3DVLM-DESS & 72.59 & 70.42 & 67.04 & 73.80 & 76.93 & 76.89 & 1,074 \\
\addlinespace[1pt]
Knee3DVLM-TSE  & 70.36 & 69.10 & 65.24 & 72.96 & 76.83 & 76.45 & 1,074 \\
\addlinespace[1pt]
Knee3DVLM      & 72.98 & 71.17 & 66.88 & 75.46 & 78.96 & 78.74 & 1,074 \\
\bottomrule
\end{tabular}}}
\end{table}

\vspace{-8pt}

\begin{table}[H]
\captionsetup{skip=3pt}
\centering
\caption{Pathology-wise balanced accuracy for the three
Knee3DVLM input configurations. Values are percentages. TSE-only and
DESS-only values were recomputed directly from their held-out
target-level prediction exports using the unweighted mean of
target-level balanced accuracy within each pathology group; Knee3DVLM
values were computed identically. Bold indicates the highest value in
each column.}\label{tab:pathology-ba}

\begin{minipage}{\textwidth}
\centering
{\fontsize{8}{9}\selectfont\setlength{\tabcolsep}{3.2pt}
\resizebox{\linewidth}{!}{%
\begin{tabular}{lrrrrrrrr}
\toprule
\textbf{Model} & \textbf{BML} & \textbf{Cartilage} & \textbf{Osteophyte} & \textbf{Meniscus} & \textbf{ACL} & \textbf{PCL\textsuperscript{*}} & \textbf{Effusion} & \textbf{Synovitis} \\
\midrule
Knee3DVLM-TSE  & 71.34 & 70.62 & 69.61 & 68.40 & \textbf{80.42} & 50.66 & \textbf{72.86} & \textbf{65.71} \\
Knee3DVLM-DESS & 69.55 & 72.77 & 71.29 & 67.18 & 75.95 & 32.51 & 71.64 & 64.22 \\
Knee3DVLM      & \textbf{71.60} & \textbf{73.48} & \textbf{71.67} & \textbf{69.02} & 78.37 & \textbf{51.60} & 72.41 & 63.60 \\
\bottomrule
\end{tabular}}}

\vspace{2pt}

\raggedright
\fontsize{7.5}{8.5}\selectfont
\textsuperscript{*}Only six positive PCL examinations were available in the held-out
cohort; the PCL estimate should therefore be interpreted with substantial
uncertainty.
\end{minipage}
\end{table}

Relative to Knee3DVLM-\allowbreak{}TSE, Knee3DVLM improved average accuracy by
2.62 percentage points, macro balanced accuracy by 2.07 points, mean
ROC-AUC by 2.13 points and macro ROC-AUC by 2.29 points.
Relative to Knee3DVLM-\allowbreak{}DESS, it improved
average accuracy by 0.39 points, macro balanced accuracy by 0.75
points, mean ROC-AUC by 2.03 points, and macro ROC-AUC by 1.85
points. Dual-sequence
fusion therefore improved every prespecified aggregate endpoint over
either single-sequence configuration.

The central finding was that model performance reflected the
complementary tissue contrasts used by radiologists in clinical
practice. Among the single-sequence configurations, Knee3DVLM-\allowbreak{}DESS
achieved higher balanced accuracy for cartilage (72.77\% vs. 70.62\%)
and osteophytes (71.29\% vs. 69.61\%), whereas Knee3DVLM-\allowbreak{}TSE performed
better for BMLs (71.34\% vs. 69.55\%), meniscal abnormalities (68.40\%
vs. 67.18\%), ACL abnormalities (80.42\% vs. 75.95\%),
effusion-synovitis (72.86\% vs. 71.64\%), and Hoffa-synovitis (65.71\%
vs. 64.22\%). Overall, the results show a clinically coherent
sequence-specific pattern rather than a uniform advantage for either
acquisition.

\begin{table}[H]
\captionsetup{skip=3pt}
\centering
\caption{Knee3DVLM performance by pathology group. Values are
percentages, computed directly from the same target-level export used
for Tables~\ref{tab:primary-binary} and~\ref{tab:pathology-ba}. Average
accuracy and macro ROC-AUC are unweighted means of per-target values
within each pathology group. Mean ROC-AUC weights each target-level
ROC-AUC by its number of evaluable examinations within the pathology
group.}\label{tab:pathology-performance}

\begin{minipage}{\textwidth}
\centering
{\fontsize{8}{9}\selectfont\setlength{\tabcolsep}{4.2pt}
\resizebox{\linewidth}{!}{%
\begin{tabular}{lrrrrr}
\toprule
\textbf{Pathology} & \textbf{Targets} & \textbf{Average accuracy} & \textbf{Balanced accuracy} & \textbf{Mean ROC-AUC} & \textbf{Macro ROC-AUC} \\
\midrule
BML & 15 & 73.05 & 71.60 & 78.88 & 78.86 \\
Cartilage & 14 & 76.01 & 73.48 & 80.45 & 80.38 \\
Meniscus & 12 & 74.31 & 69.02 & 77.58 & 78.03 \\
Osteophyte & 12 & 71.13 & 71.67 & 79.16 & 79.16 \\
ACL & 1 & 70.14 & 78.37 & 87.99 & 87.99 \\
Effusion & 1 & 67.60 & 72.41 & 77.80 & 77.80 \\
Synovitis & 1 & 63.65 & 63.60 & 70.89 & 70.89 \\
PCL\textsuperscript{*} & 1 & 53.11 & 51.60 & 57.06 & 57.06 \\
\bottomrule
\end{tabular}}}

\vspace{2pt}

\raggedright
\fontsize{7.5}{8.5}\selectfont
\textsuperscript{*}Only six positive PCL examinations were available in the held-out
cohort; the PCL estimate should therefore be interpreted with substantial
uncertainty.
\end{minipage}
\end{table}

The strongest group-level discrimination result was 87.99\% ROC-AUC for
ACL abnormalities; among the multi-target pathology groups, cartilage
achieved the highest mean ROC-AUC (80.45\%) and macro ROC-AUC (80.38\%).

Fusion provided the strongest overall model. Knee3DVLM ranked first on
every prespecified aggregate binary endpoint: average accuracy
(72.98\%), macro balanced accuracy (71.17\%), mean ROC-AUC (78.96\%),
and macro ROC-AUC (78.74\%). At
the pathology-group level, fusion achieved the highest balanced accuracy
for all four multi-target groups: BML, cartilage, osteophytes, and
meniscal abnormalities.

\subsection{Multiclass Grading and Comparison with
3DReasonKnee.}\label{multiclass-grading-and-comparison-with-3dreasonknee}

The secondary multiclass task reconstructs the exact-grade questions
used by 3DReasonKnee. We matched the released 3DReasonKnee cohort
identifiers to the original OAI examinations by subject ID, visit, and
knee side and calculated category-wise average accuracy using the metric
reported in that paper (Table~\ref{tab:3dreasonknee}). Multiclass Knee3DVLM was numerically
higher than the strongest published Qwen2.5VL-\allowbreak{}3B configuration without
chain-of-thought supervision in all five categories: BML (72.11\% vs.
70.60\%), cartilage (62.32\% vs. 60.20\%), osteophyte (35.22\% vs.
34.50\%), meniscus (40.32\% vs. 38.70\%), and Others (65.20\% vs.
61.60\%). In Table~\ref{tab:3dreasonknee}, PCL, ACL, Effusion and Synovitis are grouped under
Others solely to match the category definitions published by
3DReasonKnee. This benchmark-aligned grouping is not used in the primary
57-target analysis, in which these classes remain distinct targets.

\begin{table}[H]
\caption{Category-wise average accuracy comparison with the published
3DReasonKnee multiclass grading results. \textsuperscript{\dag}Published
results transcribed from Sambara et al.~{[}3{]}. Bold indicates the
highest value in each category.}\label{tab:3dreasonknee}

{\fontsize{8}{9.2}\selectfont\setlength{\tabcolsep}{3.8pt}
\begin{tabularx}{\textwidth}{@{}l X rrrrr@{}}
\toprule
\textbf{Model} & \textbf{Method} & \textbf{BML} & \textbf{Cartilage} & \textbf{Osteophyte} & \textbf{Meniscus} & \textbf{Others} \\
\midrule
Qwen2.5VL-3B\textsuperscript{\dag} & SFT with CoT & 66.70 & 60.00 & 34.20 & 34.00 & 59.00 \\
Med3DVLM\textsuperscript{\dag} & SFT with CoT & 67.10 & 57.90 & 33.30 & 34.30 & 60.50 \\
Med3DVLM\textsuperscript{\dag} & SFT without CoT & 67.00 & 57.80 & 33.00 & 34.60 & 60.00 \\
Qwen2.5VL-3B\textsuperscript{\dag} & SFT without CoT & 70.60 & 60.20 & 34.50 & 38.70 & 61.60 \\
\midrule
Our VLM \allowbreak{}(DESS-only) & Full training without CoT & 70.80 &
\best{65.30} & \best{36.10} & 32.90 & 64.30 \\
Our VLM \allowbreak{}(TSE-only) & Full training without CoT & 71.91 & 55.40 &
34.05 & \best{41.60} & \best{65.40} \\
Our VLM (paired DESS-TSE) & Full training without CoT & \best{72.11}
& 62.32 & 35.22 & 40.32 & 65.20 \\
\bottomrule
\end{tabularx}}
\end{table}

Values are category-wise average accuracy percentages, matching the
metric reported by 3DReasonKnee. We reconstructed the comparison cohort
by matching the released 3DReasonKnee OAI identifiers to the original
OAI data at the subject, visit, and knee-side levels. Published baseline
values are transcribed from the 3DReasonKnee paper; our values were
computed on the aligned cohort using the same category definitions.

This comparison requires an important qualification. As shown in Figure
1, Panel B, grade 0 accounted for 85.41\% of ground-truth component
labels, so exact-grade accuracy can remain high despite weak
minority-grade performance. The published 3DReasonKnee results do not
include per-grade confusion matrices; therefore, balanced accuracy
cannot be recovered for those models. Table~\ref{tab:3dreasonknee} should be interpreted as a
benchmark-aligned comparison using the authors\textquotesingle{}
reported metric rather than a complete comparison of grading
performance.

\section{Discussion}\label{discussion}

The DESS advantage for cartilage and osteophytes is consistent with its
thin-section, high-resolution three-dimensional depiction of articular
morphology and cartilage--fluid and bone--cartilage interfaces. DESS was
selected in the OAI protocol specifically to support quantitative
cartilage morphometry, and prior whole-joint studies found chondral
abnormalities well visualized with high-resolution fat-suppressed DESS
{[}5,13{]}. Osteophytes are also primarily marginal osseous structures,
making the observed DESS advantage anatomically plausible. These are
concordant associations, however, and should not be interpreted as proof
that sequence contrast alone caused the performance difference.

Conversely, the TSE advantages align with the role of fluid-sensitive
proton-density or intermediate-weighted imaging in routine knee
interpretation. Fat-suppressed TSE increases the conspicuity of marrow
edema-like signal, joint fluid, and soft-tissue edema and provides
strong contrast for meniscal and cruciate-ligament abnormalities;
current ESSR recommendations therefore emphasize multiplanar
fluid-sensitive imaging for systematic assessment of the knee {[}6{]}.
The higher TSE performance for BML, meniscus, ACL, effusion, and
synovitis is therefore clinically expected.

This pattern supports the design premise that
sequence-specific encoders preserve morphology-dominant DESS features
and fluid-sensitive TSE features before token-level fusion. A prior
multireader study similarly found that adding a short fluid-sensitive
PD-FSE acquisition to DESS improved positive and negative agreement for
comprehensive knee assessment {[}13{]}.

The secondary exact-grade experiment provided complementary benchmark
evidence. Among our models, Knee3DVLM-\allowbreak{}DESS achieved the highest
cartilage and osteophyte accuracy, Knee3DVLM-\allowbreak{}TSE achieved the highest
meniscus and Others accuracy, and Knee3DVLM achieved the highest BML
accuracy. More importantly for the external comparison, Knee3DVLM was
numerically higher than the strongest published 3DReasonKnee
configuration in every shared category: BML (72.11\% vs. 70.60\%),
cartilage (62.32\% vs. 60.20\%), osteophyte (35.22\% vs. 34.50\%),
meniscus (40.32\% vs. 38.70\%), and Others (65.20\% vs. 61.60\%).
Finally, this study evaluates diagnostic targets and deterministic
structured reports rather than prospective reader performance, workflow
impact, or patient outcomes.

\section{Conclusion}\label{conclusion}

Knee3DVLM addresses a gap in three-dimensional knee MRI by combining
full-volume, whole-joint assessment with paired DESS and fluid-sensitive
TSE inputs. On the primary binary task, the fused model achieved the
highest average accuracy, balanced accuracy, mean ROC-AUC, and
macro ROC-AUC among the three tested configurations.
On the secondary multiclass exact-grade benchmark, Knee3DVLM was
numerically higher than the strongest published 3DReasonKnee baseline in
all five reported categories. These results support sequence-aware
fusion for comprehensive knee MRI assessment while underscoring the need
for class-aware evaluation.

\section*{Acknowledgments}
This study was supported by a Cleveland Clinic Catalyst Grant.

\clearpage
\begingroup
\thispagestyle{plain}
\renewcommand{\refname}{References}
\makeatletter
\renewcommand{\@biblabel}[1]{\textbf{[#1]}}
\makeatother
\fontsize{10.5}{11.5}\selectfont
\setlength{\parskip}{0pt}
\setlength{\emergencystretch}{3em}

\endgroup


\begin{thebibliography}{13}
\setlength{\itemsep}{1pt}
\setlength{\parsep}{0pt}
\setlength{\topsep}{0pt}
\setlength{\partopsep}{0pt}
\setlength{\labelsep}{0.6em}
\setlength{\leftmargin}{2.1em}
\setlength{\itemindent}{0pt}
\setlength{\parfillskip}{0pt plus 1fil}
\raggedright

\bibitem{ref1}
F. Bai, Y. Du, T. Huang, M. Q.-H. Meng, and B. Zhao, ``M3D:
Advancing 3D Medical Image Analysis with Multi-Modal Large Language
Models,'' arXiv:2404.00578, 2024.
\url{https://arxiv.org/abs/2404.00578}

\bibitem{ref2}
Y. Xin, G. C. Ates, K. Gong, and W. Shao, ``Med3DVLM: An
Efficient Vision-Language Model for 3D Medical Image Analysis,''
arXiv:2503.20047, 2025.
\url{https://arxiv.org/abs/2503.20047}

\bibitem{ref3}
S. Sambara, S. E. Kim, X. Zhang, L. Luo, S. Johri, M. Baharoon,
D. H. Ro, and P. Rajpurkar, ``3DReasonKnee: Advancing Grounded Reasoning
in Medical Vision Language Models,'' Pacific Symposium on Biocomputing,
vol. 31, pp. 99--113, 2026.
\url{https://doi.org/10.1142/9789819824755_0008}

\bibitem{ref4}
M. C. Nevitt, D. T. Felson, and G. Lester, ``The Osteoarthritis
Initiative: Protocol for the Cohort Study,'' Osteoarthritis Initiative,
version 1.1, 2006.
\url{https://nda.nih.gov/static/docs/StudyDesignProtocolAndAppendices.pdf}

\bibitem{ref5}
C. G. Peterfy, E. Schneider, and M. Nevitt, ``The Osteoarthritis
Initiative: Report on the Design Rationale for the Magnetic Resonance
Imaging Protocol for the Knee,'' Osteoarthritis and Cartilage, vol. 16,
no. 12, pp. 1433--1441, 2008.
\url{https://doi.org/10.1016/j.joca.2008.06.016}

\bibitem{ref6}
A. P. Parkar and M. E. A. P. M. Adriaensen, ``ESR Essentials:
MRI of the Knee---Practice Recommendations by ESSR,'' European
Radiology, vol. 34, no. 10, pp. 6590--6599, 2024.
\url{https://doi.org/10.1007/s00330-024-10706-7}

\bibitem{ref7}
F. Eckstein, M. Hudelmaier, W. Wirth, B. Kiefer, R. Jackson, J.
Yu, C. B. Eaton, and E. Schneider, ``Double Echo Steady State Magnetic
Resonance Imaging of Knee Articular Cartilage at 3 Tesla: A Pilot Study
for the Osteoarthritis Initiative,'' Annals of the Rheumatic Diseases,
vol. 65, no. 4, pp. 433--441, 2006.
\url{https://doi.org/10.1136/ard.2005.039370}

\bibitem{ref8}
A. S. Chaudhari, M. S. Black, S. Eijgenraam, W. Wirth, S.
Maschek, B. Sveinsson, F. Eckstein, E. H. G. Oei, G. E. Gold, and B. A.
Hargreaves, ``Five-Minute Knee MRI for Simultaneous Morphometry and T2
Relaxometry of Cartilage and Meniscus and for Semiquantitative
Radiological Assessment Using Double-Echo in Steady-State at 3T,''
Journal of Magnetic Resonance Imaging, vol. 47, no. 5, pp. 1328--1341,
2018.
\url{https://doi.org/10.1002/jmri.25883}

\bibitem{ref9}
M. Zhang, J. B. Driban, L. L. Price, G. H. Lo, and T. E.
McAlindon, ``Magnetic Resonance Image Sequence Influences the
Relationship Between Bone Marrow Lesion Volume and Pain: Data from the
Osteoarthritis Initiative,'' BioMed Research International, vol. 2015,
article 731903, 2015.
\url{https://doi.org/10.1155/2015/731903}

\bibitem{ref10}
C. Li, C. Wong, S. Zhang, N. Usuyama, H. Liu, J. Yang, T.
Naumann, H. Poon, and J. Gao, ``LLaVA-Med: Training a Large
Language-and-Vision Assistant for Biomedicine in One Day,''
arXiv:2306.00890, 2023.
\url{https://arxiv.org/abs/2306.00890}

\bibitem{ref11}
M. Moor, Q. Huang, S. Wu, M. Yasunaga, C. Zakka, Y. Dalmia, E.
P. Reis, P. Rajpurkar, and J. Leskovec, ``Med-Flamingo: A Multimodal
Medical Few-Shot Learner,'' arXiv:2307.15189, 2023.
\url{https://arxiv.org/abs/2307.15189}

\bibitem{ref12}
D. J. Hunter, A. Guermazi, G. H. Lo, A. J. Grainger, P. G.
Conaghan, R. M. Boudreau, and F. W. Roemer, ``Evolution of
Semi-Quantitative Whole-Joint Assessment of Knee OA: MOAKS (MRI
Osteoarthritis Knee Score),'' Osteoarthritis and Cartilage, vol. 19, no.
8, pp. 990--1002, 2011.
\url{https://doi.org/10.1016/j.joca.2011.05.004}

\bibitem{ref13}
A. S. Chaudhari, K. J. Stevens, B. Sveinsson, J. P. Wood, C. F.
Beaulieu, E. H. G. Oei, J. K. Rosenberg, F. Kogan, M. T. Alley, G. E.
Gold, and B. A. Hargreaves, ``Combined 5-Minute Double-Echo in
Steady-State With Separated Echoes and 2-Minute Proton-Density-Weighted
2D FSE Sequence for Comprehensive Whole-Joint Knee MRI Assessment,''
Journal of Magnetic Resonance Imaging, vol. 49, no. 7, pp. e183--e194,
2019.
\url{https://doi.org/10.1002/jmri.26582}

\end{thebibliography}
\end{document}